\documentclass{article}
\usepackage{array}

\usepackage{PRIMEarxiv}
\usepackage{amsmath}
\usepackage[utf8]{inputenc} 
\usepackage[T1]{fontenc}    
\usepackage{hyperref}       
\usepackage{url}            
\usepackage{booktabs}       
\usepackage{amsfonts}       
\usepackage{nicefrac}       
\usepackage{microtype}      
\usepackage{lipsum}
\usepackage{fancyhdr}       
\usepackage{graphicx}       
\usepackage{subcaption}
\usepackage{afterpage}
\usepackage[linesnumbered,ruled,vlined]{algorithm2e}
\SetCommentSty{textnormal}
\usepackage{algpseudocode}
\usepackage{multirow}
\usepackage{float}
\usepackage{multicol}
\usepackage{array}
\usepackage{makecell}
\graphicspath{{media/}}     
\usepackage[colorinlistoftodos,prependcaption,textsize=tiny]{todonotes}
\usepackage{cleveref}
\usepackage[toc,page]{appendix}

\usepackage[normalem]{ulem}
\newcommand{\zhou}[1]{{\color{black}{#1}}} 

\newcommand{\chen}[1]{{\color{black}{#1}}}

\title{Unifying Sequences, Structures, and Descriptions \\ for Any-to-Any Protein Generation with \\ the Large Multimodal Model HelixProtX}

\author{
  Zhiyuan Chen\thanks{Equal contributions.}, Tianhao Chen\footnotemark[1], Chenggang Xie, Yang Xue, Xiaonan Zhang, Jingbo Zhou\footnotemark[2], Xiaomin Fang\thanks{Corresponding author: Xiaomin Fang (fangxiaomin01@baidu.com) and Jingbo Zhou (zhoujingbo@baidu.com).} \\
  PaddleHelix Team \\
  Baidu Inc. \\
  \texttt{\url{https://paddlehelix.baidu.com/}} \\
  \today \\
}

\begin{document}
\maketitle

\begin{abstract}

\zhou{Proteins are fundamental components of biological systems and can be represented through various modalities, including sequences, structures, and textual descriptions. Despite the advances in deep learning and scientific large language models (LLMs) for protein research, current methodologies predominantly focus on limited specialized tasks -- often predicting one protein modality from another. These approaches restrict the understanding and generation of multimodal protein data. In contrast, large multimodal models have demonstrated potential capabilities in generating any-to-any content like text, images, and videos, thus enriching user interactions across various domains. Integrating these multimodal model technologies into protein research offers significant promise by potentially transforming how proteins are studied. To this end, we introduce HelixProtX, a system built upon the large multimodal model, aiming to offer a comprehensive solution to protein research by supporting any-to-any protein modality generation. Unlike existing methods, it allows for the transformation of any input protein modality into any desired protein modality.  The experimental results affirm the advanced capabilities of HelixProtX, not only in generating functional descriptions from amino acid sequences but also in executing critical tasks such as designing protein sequences and structures from textual descriptions. Preliminary findings indicate that HelixProtX consistently achieves superior accuracy across a range of protein-related tasks, outperforming existing state-of-the-art models. By integrating multimodal large models into protein research, HelixProtX opens new avenues for understanding protein biology, thereby promising to accelerate scientific discovery.}

\end{abstract}

\keywords{Protein generation \and Any-to-any generation \and Multiple modalities \and Large language model}

\section{Introduction}

\begin{figure*}[ht!]
\center
\includegraphics[width=0.8\textwidth]{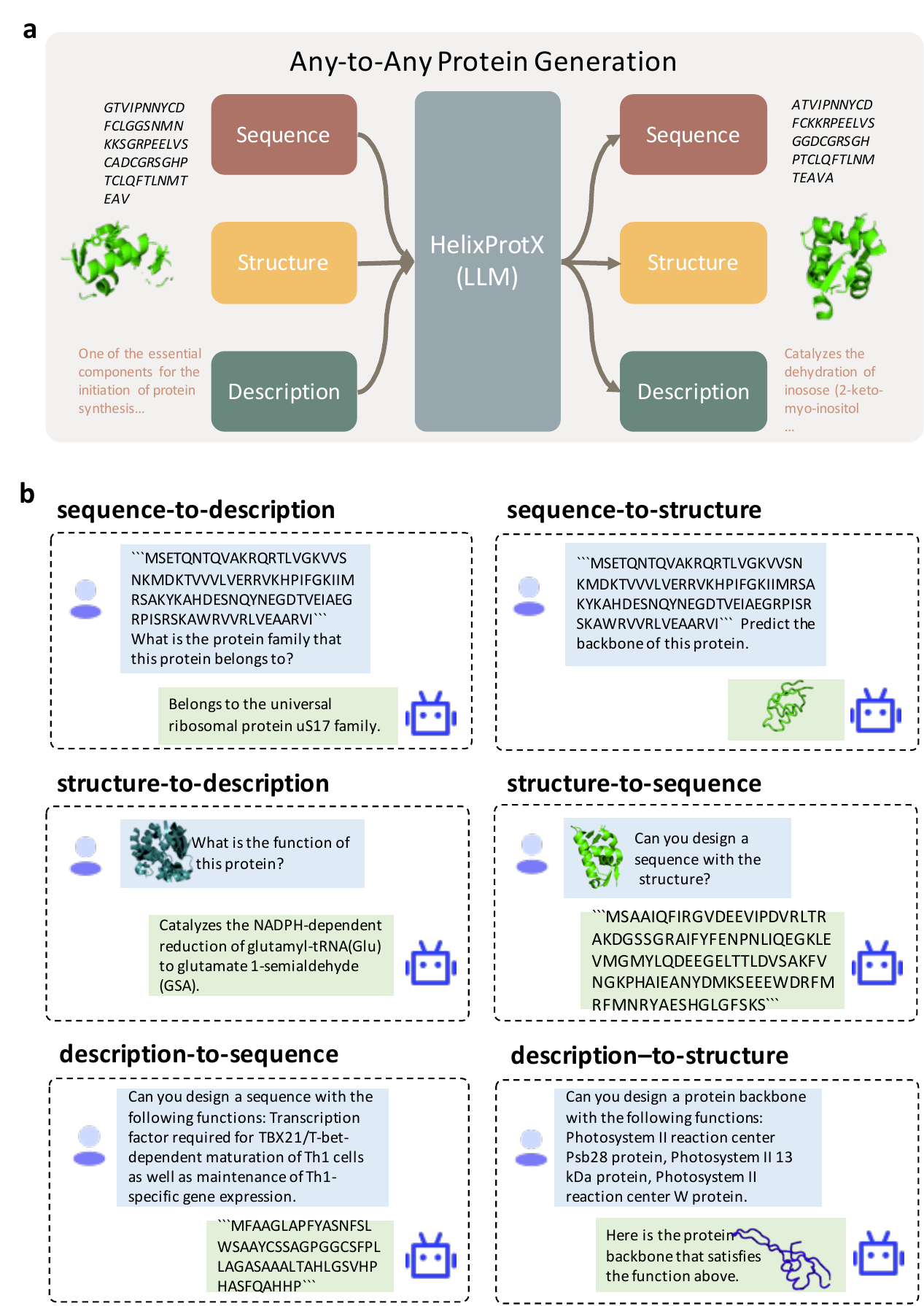}
\caption{\textbf{Overview of HelixProtX.} \textbf{a} Any-to-any protein generation produces diverse output modalities (sequence, structure, and description) from any input modality (sequence, structure, or description). 
\textbf{b} Large language model (LLM)-based system for protein any-to-any protein generation.}
\label{fig:overview}
\end{figure*}

Proteins are \zhou{fundamental} entities in the life sciences, performing diverse and crucial functions within living organisms. They provide structural support, catalyze biochemical reactions as enzymes, and facilitate the transport and storage of essential molecules. \zhou{Additionally,} proteins also \zhou{play a critical role in} a wide array of biological functions, involved in cell signaling, immune response, and gene regulation.
\zhou{The study of proteins can be approached through multiple modalities including amino acid sequences, three-dimensional structures, and textual descriptions -- each offering unique insights into understanding proteins from different perspectives:}
(1) The amino acid sequence, \zhou{also} referred to as the primary structure of a protein, encodes the genetic information and is \zhou{typically} analyzed \zhou{using} sequence-based deep learning models, \zhou{such as those outlined by} \cite{hashemifar2018predicting} to investigate the associations across the residues. (2) The \zhou{three-dimensional} structural conformation, \zhou{also known as the} folded state, \zhou{greatly} influences a protein's functional activities. Various \zhou{advanced} structural encoders, \zhou{like Geometric Vector Perceptron (GVP)} \cite{jing2020learning}, \zhou{Protein Message-Passing Neural Network (ProteinMPNN)} \cite{prompnn}, and \zhou{Invariant Point Attention (IPA)}\cite{jumper2021highly}, have been \zhou{investigated to elucidate the complex} spatial interactions among residues or atoms within a protein. (3) \zhou{The textual description offers a narrative perspective on protein functionalities as documented in the scientific literature. This textual modality provides valuable insights that enhance our understanding of the diverse roles and characteristics of proteins.}

\zhou{Deep learning-based protein research predominantly focuses on singular-task methodologies, with most studies aiming to predict one protein modality from another.  For example, many studies, such as  \cite{liu2024prott3,wang2024protchatgpt,gligorijevic2021structure,zhang2023protein}, have inferred biological protein functions from sequences \cite{liu2024prott3, wang2024protchatgpt} or structures \cite{gligorijevic2021structure, zhang2023protein}. This process, which facilitates functional annotation and the discovery of novel functions, essentially generates textual descriptions from protein sequence or structure data (sequence-to-description and structure-to-description).  Recently, deep learning has significantly advanced protein structure prediction, as demonstrated by AlphaFold II \cite{jumper2021highly} and RoseTTAfold \cite{doi:10.1126/science.abj8754}, which predicts the three-dimensional protein structural conformation from amino acid sequences (sequence-to-structure). Furthermore, AI-driven protein design has garnered considerable research interest. This field aims to generate sequences that can fold into specific three-dimensional structures (structure-to-sequence) \cite{qiu2024instructplm, prompnn, 10.5555/3618408.3620189} and to create protein sequences \cite{madani2023large, kucera2022conditional, yuan2024functional, singh2024chroma} or structures \cite{singh2024chroma} based on a given tag \cite{madani2023large, kucera2022conditional} or description \cite{yuan2024functional, singh2024chroma} (description-to-sequence or description-to-structure). Despite significant technological advancements in these tasks, current models typically specialize in singular protein-related tasks. This specialization requires researchers to understand multiple models to meet their objectives, thereby complicating the research process. Moreover, the need to manage varying input and output formats and reconcile potentially inconsistent findings across different models further increases the research burden. A holistic solution capable of addressing multiple protein-related challenges simultaneously would substantially reduce these complexities and streamline research efforts.}

\zhou{Ongoing efforts have been made to develop specialized scientific Large Language Models (LLMs) for the life sciences, yet these primarily focus on one or two protein-related tasks. Given the significant success of LLMs in addressing a wide range of tasks across diverse fields \cite{chang2024survey}, the introduction of scientific LLMs for protein research represents a highly promising and intriguing direction. However, current efforts remain somewhat constrained, primarily focusing on processing natural language queries or converting text descriptions into other modalities. For instance, BioMedGPT \cite{luo2023biomedgpt} excels in responding to natural language inquiries in bioinformatics, small molecule analysis, and protein research. Meanwhile, MolInstruction \cite{fang2023molinstructions} is capable of predicting molecular functions and using textual descriptions to design protein sequences within language models.}

\zhou{
Nevertheless, our perspective is that integrating the large multimodal model with any-to-any generation capabilities into protein research holds considerable potential to transform how proteins are studied.  Recent advancements in large multimodal model research \cite{wu2023multimodal}, particularly the any-to-any generation technique \cite{tang2023anytoany,tang2023codi2,anonymous2023nextgpt} which integrates multiple tasks into a unified model, demonstrate the potential for high-quality content generation across modalities such as text, images, speech, and video. For example, CoDi, introduced in \cite{tang2023anytoany}, utilizes diffusion models to facilitate any-to-any multimodal generation. This capability is further enhanced in CoDi-2 \cite{tang2023codi2} and NExT-GPT \cite{wu2023next}, both of which exploit the synergy between LLMs and diffusion models to improve any-to-any multimodal generation performance. Despite the considerable potential, the application of any-to-any multimodal generation techniques to protein research remains largely unexplored, offering an exciting avenue for future investigations.}


\zhou{To address this potential, we introduce HelixProtX, a system built upon a large multimodal model for any-to-any protein generation. Designed to unify protein sequences, structures, and descriptions, HelixProtX offers a comprehensive solution for protein research by facilitating any-to-any protein generation. The primary goal of HelixProtX is to leverage any-to-any protein generation to transform the methodologies used in protein research, thereby making protein-related investigation more efficient and comprehensive. As depicted in Figure~\ref{fig:overview}a, HelixProtX enables the generation of different protein modalities from any given input by seamlessly integrating sequences, structures, and textual descriptions. Upon user interaction, the system analyzes the semantics and context of each query before generating precise and detailed responses across different protein modalities, as illustrated in Figure~\ref{fig:overview}b.}


\zhou{We systematically evaluated the performance of HelixProtX across a diverse array of protein-related tasks, including description prediction (sequence-to-description and structure-to-description), sequence design (structure-to-sequence and description-to-sequence), and structure prediction/design (sequence-to-structure and description-to-structure), to verify the effectiveness of HelixProtX. Across most tasks, HelixProtX consistently outperformed existing benchmarks, leveraging the advanced capabilities of large multimodal models. The system demonstrated robustness across proteins of varying lengths and families. Notably, HelixProtX exhibited exceptional potential in text-guided protein design applications, including description-to-sequence and description-to-structure tasks. The proteins designed by HelixProtX closely matched reference proteins in terms of compatibility, with the distributions of the designed protein sequences and structures showing rationality and coherence. By harnessing the capabilities of large multimodal models for any-to-any protein generation, HelixProtX can provide novel insights into protein biology and has the potential to significantly accelerate scientific discovery in this field.}

\section{Method}
\begin{figure*}[t]
\center
\includegraphics[width=1.0\textwidth]{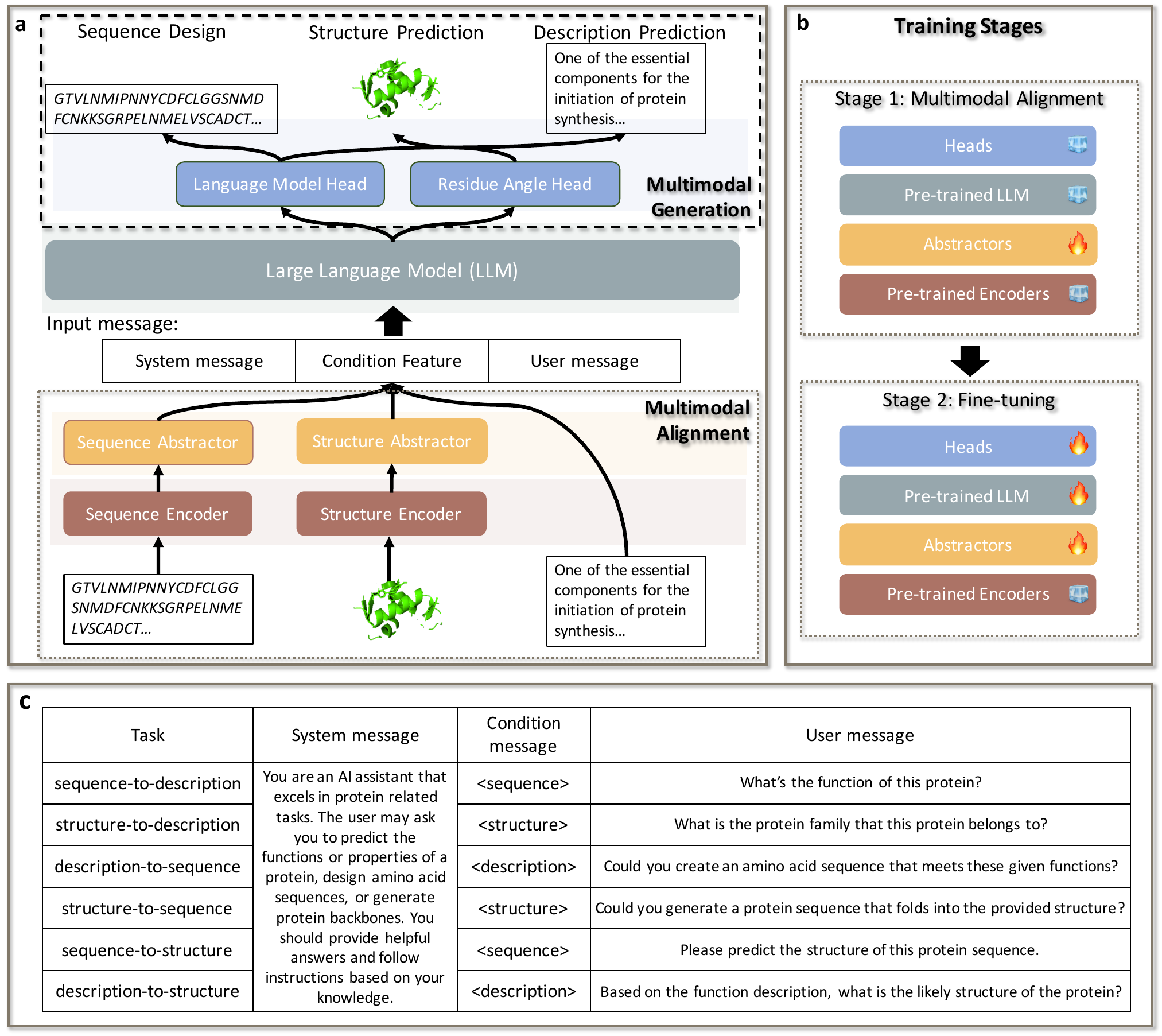}
\caption{\textbf{HelixProtX model architecture, training paradigm, and data overview.} a. The architecture of multimodal language model. b. Training paradigm. c. Demonstration of the input format of various protein-related tasks.}
\label{fig:architecture}
\end{figure*}
\subsection{Overview}
HelixProtX is an advanced multi-modal system \zhou{designed} to comprehend and generate diverse protein modalities \zhou{including sequence, structure, and description}. It facilitates user interaction through natural language-style question answering, \zhou{addressing a variety of protein-related tasks.} 

The architecture of HelixProtX, \zhou{as} illustrated in Figure~\ref{fig:architecture}a, is centered on the large language model LLM, utilizing ERNIE-Lite, a streamlined version of ERNIE BOT\footnote{https://yiyan.baidu.com/welcome}. 
\zhou{This model processes inputs composed of three key components:} the system message, the condition message, and the user message, as shown in Figure~\ref{fig:architecture}c. The system message provides contextual cues to guide the LLM's responses across various generation tasks, the condition message specifies the generation conditions, and the user message \zhou{delivers} specific input from the user.

Building on the foundational architecture centered around the LLM, HelixProtX further enhances its capabilities by aligning multimodalities of proteins. A pre-trained Sequence Encoder and a pre-trained Structure Encoder are introduced to encode amino acid sequences and the three-dimensional structures, respectively. To ensure alignment among different protein modalities and textual queries, we incorporate Sequence Abstractor and Structure Abstractor modules. These modules condense sequence and structure information into a set of learnable tokens. For queries that depend on protein sequences or structures as conditions, their representations are generated by abstractors and serve as the condition message. For queries dependent on descriptions as conditions, the corresponding text is used directly as the condition message.

\zhou{HelixProtX supports tasks across multiple protein modalities, with a primary focus on predicting functional descriptions, designing protein sequences, and predicting or designing protein structures.} In tasks like description prediction and sequence design, the Language Model Head leverages the latent representations \zhou{from the LLM to generate responses by computing probability distributions across token vocabularies.} 
For the task of structure prediction/design, we introduce a specialized Residue Angle Head, designed to forecast the six angles describing the relative position of all backbone atoms in the residue, thereby \zhou{facilitating} the \zhou{prediction} of the protein backbone structure.

The training process of HelixProtX is divided into two stages, as depicted in Figure~\ref{fig:architecture}b. Initially, the parameters of the abstractors are optimized exclusively during the first stage. Subsequently, in the second stage, gradients are backpropagated to optimize all \zhou{all model parameters.} 

\subsection{Input Format}
The input of HelixProtX generally follows the conventional paradigm of multimodal instruction tuning \cite{zhang2023instruction}. Each input consists of three parts: a system message, a condition message, and a user message. The input format for various protein-related tasks is illustrated in Figure~\ref{fig:architecture}c.

\zhou{ The system message provides an overview, helping the model understand its role and capabilities within the system.}
\zhou{It} specifies the model's characteristics and responsibilities, serving as a comprehensive guide to ensure that the model resolves protein-related tasks.

Following the system message, the condition message provides the specific conditions required to perform the given task. For example, in a sequence design task, the condition message might include details of the desired protein structure and function description. \zhou{In the HelixProtX model,} the condition message could be the output of a sequence abstractor, a structure abstractor, or a representation of a functional description.

\zhou{Lastly, the user message clarifies the intentions and specific requests of the user. It provides essential context and details, enabling HelixProtX to accurately perform specialized tasks. }
The user message serves to articulate the user intent, instructing HelixProtX on which task to perform. \zhou{Examples of user messages for different tasks can be found in Figure~\ref{fig:architecture}c. }

\subsection{Multimodal Encoders with Alignment} 

\zhou{Conventional natural language models\cite{min2023recent} typically only accept natural language tokens as input. To enable their capability to handle both amino acid sequence and three-dimensional structural modalities, we introduce two pre-trained modules: the Sequence Encoder and the Structure Encoder. The Sequence Encoder is designed to capture representations of amino acid sequences, whereas the Structure Encoder focuses on structural modalities. While it is possible to directly input amino acid sequences and their coordinates into a natural language model without any encoding, this approach may fail to effectively capture the important co-evolutionary and spatial relationships between amino acids. These relationships are crucial for accurately representing the complexity and multifaceted nature of proteins. To address this, we introduce these two specialized encoders, enhancing the model's understanding and processing of protein data.}

\textbf{Sequence Encoder.}  \zhou{The amino acid sequence of a protein serves as the input for the sequence encoder.}
We opt to employ our previous work, HelixFold-Single \cite{fang2023method}, as the Sequence Encoder, \zhou{chosen for its robust protein structure prediction capabilities}. HelixFold-Single is composed of a Protein Language Model (PLM) and a Geometry Modeling module. The PLM, \zhou{undergoing} self-supervised learning on a comprehensive protein sequence dataset, \zhou{can capture} co-evolutionary information\cite{carbone2011co} among amino acids. \zhou{Subsequently,} \zhou{the Geometry Modeling} module extracts spatial relationships among them.
\zhou{Specifically, to process each amino acid sequence, it is input into HelixFold-Single. The sequence first is processed by the PLM to understand co-evolutionary patterns. Following this, the representation is then fed into the Geometry Modeling module to refine this data into a spatial context.  The output from this module is then utilized as the input for the Abstractor.}


\textbf{Structure Encoder.}
\zhou{The three-dimensional structure of a protein serves as the input for the Structure Encoder. } \zhou{We utilize} the well-established protein design model, ProteinMPNN \cite{prompnn}, \zhou{as the Structure Encode}. Specifically, we used the output representations from the last encoder layer \zhou{of ProteinMPNN} as structure embedding. \zhou{ProteinMPNN has been extensively trained on} tens of thousands of protein structures curated from the Protein Data Bank (PDB). Through this rigorous training process, ProteinMPNN has acquired a deep understanding of the intricate spatial interactions among backbone atoms within proteins. This comprehensive knowledge enables ProteinMPNN to effectively encode structural information. \zhou{To this end, the structure representation derived from the final encoder layer thus serves as an input for the Abstractor within HelixProtX.}

\textbf{Abstractors.} 
\zhou{We further introduce an Abstractor module to conduct the multimodal alignment, essential for integrating the diverse representations generated by the encoders discussed above. } The Sequence Encoder and Structure Encoder produce Sequence and Structure representations, capturing sequence and structural information in their respective feature spaces. However, these \zhou{representations} are inherently distant from the natural language feature space of the LLM, making it challenging for the LLM to effectively utilize them directly. To bridge this gap, we introduce an Abstractor module, following the approach in \cite{ye2023mplugowl}, which incorporates several learnable tokens that interact with the input Sequence and Structure representations, \zhou{adjusting and aligning them to fit the LLM's feature space}. This interaction \zhou{facilitates} the integration of both Sequence and Structure representations into the LLM's feature space, \zhou{enabling it to effectively leverage this multimodal information. The design details of the Abstractor module are provided in Appendix Section \ref{subsec:abstractor}.}

\subsection{Multimodal Generation}
\zhou{In this section, we introduce the implementation of multimodal generation within HelixProtX. Unlike typical LLMs that decode only sequential information through a Language Model Head, HelixProtX is uniquely equipped to handle both sequence and structural decoding, broadening its applicability in processing protein data. For sequence decoding, we employ the standard Language Model Head, which generates sequential data of proteins. Complementarily, structural decoding utilizes a novel Residue Angle Head, specifically designed to predict the angles between amino acids. This component enables the precise reconstruction of a protein’s three-dimensional structure.}

\textbf{Language Model Head to Learn Description Prediction and Sequence Design.}
In the realm of generative pre-trained models (e.g. GPT \cite{yenduri2024gpt} ), a commonly adopted approach for pre-training involves maximizing the likelihood of  predicting the subsequent token in a sequence based on its preceding tokens. This is achieved through the use of the negative log-likelihood (NLL) loss function during \zhou{a} training phase. For tasks encompassing textual description generation and amino acid sequence generation, we adhere to this prevalent pre-training loss mechanism to effectively fine-tune the model for improved performance and adaptability. The Language Model Head and the NLL loss function $L_{\text{NLL}}(\mathbf{x})$ are defined as:
\begin{align} 
\mathcal{P}(x_i | x_{<i}) &= \text{Softmax}(\text{LanguageModelHead}(h_i)), \\
L_{\text{NLL}}(\mathbf{x}) &= -\sum_{i=j}^{n} \log(\mathcal{P}(x_i | x_{<i})),
\end{align}
where LanguageModelHead(.) is a \zhou{multi-layer perceptron.} 
$x_i$ represents the $i$-th token, $h_i$ denotes the representation of token $x_i$ produced by the final layer of the Language Model, $j$ is the starting index of the response, and $n$ is the number of input tokens.

\textbf{Residue Angle Head to Learn Structure Prediction/Design.}
The structure of a protein can be characterized by a sequence of angles that describe the relative orientation of its backbone atoms. In a manner similar to the decoding process of conventional language models, HelixProtX adopts an autoregressive approach to decode the angles associated with each amino acid. Following prior research \cite{wu2024protein}, HelixProtX predicts the six angles for amino acids rather than the coordinates of atoms with variable numbers, streamlining the structure decoding process and preserving the inherent translational and rotational invariance of the protein backbone structure. Once the Language Model Head decodes a special token [NUM], the Residue Angle Head is activated to output the angles of the amino acids. The outputted angles are adapted into the word embedding space as the input of the LLM to produce the next token. \zhou{Details on how to generate such Residue Angle Embedding are provided in the Appendix, Section \ref{subsec:angle_emb}, which also offers an in-depth understanding of this process.}

The Residue Angle Head and the corresponding angle loss function $L_\text{angle}(\mathbf{\hat{y}}, \mathbf{y})$ are defined as:
\begin{align}
    \hat{y}_i &= \text{ResidueAngleHead}(h_i), \qquad \hat{y}_i \in \mathbb{R}^6\\
    L_\text{angle}(\mathbf{\hat{y}}, \mathbf{y}) &= \sum_{i=1}^{n} (L_\text{distance}(\hat{y}_i, y_i) + \lambda L_\text{constraint}(\hat{y}_i)),\\
    L_\text{distance}(\hat{y}_i,y_i) &= \frac{1}{6}\sum_{j=1}^6|((\hat{y}_{ij} - y_{ij}) + \pi) \bmod (2\pi) - \pi|, \\
    L_\text{constraint}(\hat{y}_i) &= \frac{1}{6}\sum_{j=1}^6\max(0, -(\pi+\epsilon) - \hat{y}_{ij}, \hat{y}_{ij}-(\pi+\epsilon)).
\end{align}


Here, ResidueAngleHead(.) is a multiple-layer perception. \chen{The total number of amino acids is represented by $n$, with $\mathbf{\hat{y}}$ and $\mathbf{y}$ respectively denoting the predicted and actual angles for all amino acids.} $\hat{y}_i$ denotes the predicted six angles of the $i$-th amino acid, and the ground-truth angles are denoted as $y_{i}$. We minimize the distance between predicted angles and their ground-truth counterparts through 
\zhou{error function $L_{distance}(.)$}, while ensuring that predicted angles adhere to the range $-\pi$ to $\pi$ through 
\zhou{penalty function $L_{constraint}(.)$} \chen{, which incorporates a hyperparameter $\epsilon$ that defines the allowable range of deviation for $\hat{y}_i$ from the interval $[-\pi, \pi]$ during training}. The hyperparameter $\lambda$ is introduced to balance these two functions.

\subsection{Training Stages}

\zhou{HelixProtX employs a two-stage training approach to optimize the model. The first stage focuses on optimizing the Abstractor module while other modules remain frozen, aiming to align the sequence, structure, and description representations. This alignment is crucial for ensuring that the model comprehensively integrates these varied modalities of protein data. The second stage is dedicated to addressing specific protein-related tasks, during which all modules are optimized except for the pre-trained encoders. The detailed architecture and training process are illustrated in Figure~\ref{fig:architecture}b, which visually outlines the progression and focus of each training stage.}

\zhou{\textbf{Stage I.}} In the first stage, we align the sequence representation \zhou{proudced} by the Sequence Encoder and the structure representation \zhou{proudced}  by the Structure Encoder to the textual description representation. All model parameters are kept frozen except for those of the Sequence Abstractor and the Structure Abstractor. We use data from sequence-to-description and structure-to-description tasks for training, ensuring the \zhou{effective} alignment of sequence and structure representations with the natural language feature space of the language model.


\zhou{\textbf{Stage II.}} \zhou{In the second stage, we focus on training for protein-related tasks, as depicted in Figure~\ref{fig:architecture}b. During this stage, we optimize all model parameters except those of the Sequence Encoder and Structure Encoder, which remain fixed to preserve the integrity of their initial representations. To ensure balanced learning across various tasks, we sample data for each of the six protein-related tasks with a probability of 15\%.  This approach allows the model to evenly develop capabilities to tackle these protein-related challenges.  Additionally, we sample dialogue data with a 10\% probability to maintain the model's conversational abilities. This balanced sampling strategy is crucial for equipping the model to effectively address both specialized and general interaction tasks.}

\section{Results}
We assess the potential of HelixProt in protein-related tasks by focusing on three key areas: description prediction, sequence design, and structure prediction/design. \zhou{Furthermore, we demonstrate the benefits of training a unified model--capable of handling multiple protein tasks—over the traditional approach of training multiple task-specific models independently }

\zhou{To build a comprehensive multimodal dataset, we curated resources from UniProtQA \cite{luo2023biomedgpt}, which provided functional descriptions and sequence data, and SwissProt \cite{boeckmann2003swiss}, which contributed structural data. We organized each data item into six pairings, creating modal pairs that support different multimodal protein tasks including sequence-to-description, structure-to-description, description-to-sequence, structure-to-sequence, sequence-to-structure, and description-to-structure. This approach yielded a dataset of $361,498 \times 6 = 2,168,988$ instances. We have divided the datasets for each task into training, validation, and test sets, maintaining an 8:1:1 ratio to ensure training and evaluation of the model’s performance.}

\subsection{Description Prediction}
\begin{figure*}[h]
\center
\includegraphics[width=1.0\textwidth]{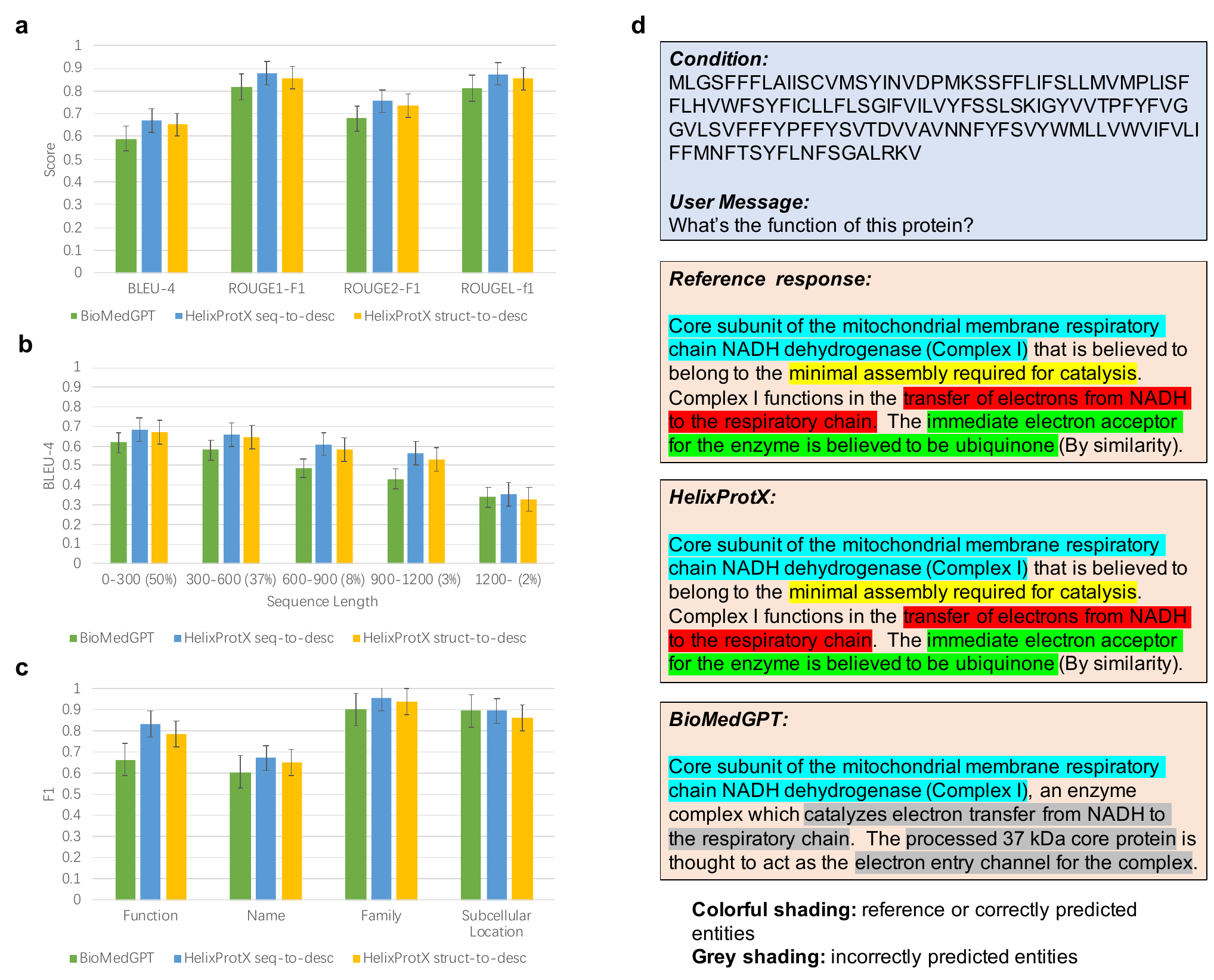}
\caption{\textbf{Results of description prediction.} a. Overall text coherence comparison. b. Impact of protein sequence length. c. Protein function accuracy comparison. d. A case illustrating description prediction task.}
\label{fig:description_prediction}
\end{figure*}

The description prediction task involves \zhou{generating} the functional description of a protein from its sequence or structure. A protein's functional description is encapsulated in a series of natural language texts that pertain to four distinct categories: the protein's function, its classification within a protein family, its official name, and its subcellular localization.
To investigate HelixProtX's potential for protein \zhou{description prediction}, we present the results of HelixProtX in inferring functional descriptions from protein sequences (denoted as HelixProtX seq-to-desc) and from protein structures (denoted as HelixProtX struct-to-desc) in Figure~\ref{fig:description_prediction}. 
\zhou{For the sequence-to-description comparison,} we use BioMedGPT \cite{luo2023biomedgpt}, another bioinformatics large language model, as the baseline. 
\zhou{For the structure-to-description task, to our knowledge, there are no existing large models that effectively address this task.}

Firstly, we \zhou{begin by analyzing} the overall performance of HelixProtX in \zhou{predicting functional descriptions}. We employ commonly used evaluation metrics in natural language processing, namely BLEU score and ROUGE score \cite{lin2004rouge}, which are \zhou{usually} used in machine translation and summarization tasks, to assess the consistency between generated functional descriptions and ground truth descriptions. As shown in Figure ~\ref{fig:description_prediction}a, HelixProtX outperforms BioMedGPT across all metrics in the sequence-to-description task, indicating higher accuracy and fluency in the generated descriptions. \zhou{Although there is no direct baseline for the structure-to-function task}, comparing HelixProtX struct-to-desc with BioMedGPT allows us to infer that HelixProtX performs well in this task as well. Compared to HelixProtX seq-to-desc, the scores for HelixProtX struct-to-desc are slightly lower. We speculate that this could be attributed to the advantage of HelixProtX's sequence encoder, HelixFold-Single, \zhou{which benefits from a greater} quantity and diversity of training data \zhou{encompassing} both sequence and structural information, \zhou{compared to} its structure encoder ProteinMPNN.
We further examined the performance of the models across different protein sequence lengths, as illustrated in Figure ~\ref{fig:description_prediction}b. As the length of the protein sequences increased, all models showed a decreasing trend in BLEU-4 scores. This decline \zhou{likely reflects the increasing complexity and variability of functions in longer protein sequences}, making it more difficult to accurately infer complete functional descriptions. Notably, HelixProtX consistently outperformed the baseline method across all protein sequence length intervals, highlighting \zhou{its robustness in handling complex protein data.}

\zhou{We further analyzed the accuracy of predicted protein functions to demonstrate the advantages of HelixProtX.}
Leveraging the advanced named entity recognition capabilities of \chen{LLM}, we extract functional entities from both reference descriptions and model predictions, as illustrated in Figure~\ref{fig:description_prediction}d. Then, we evaluate their alignment using the F1 score, which offers a balanced evaluation by considering both the precision and recall of predicted entities compared to ground truth. Figure~\ref{fig:description_prediction}c details results across four description groups: functionality, official name, family, and subcellular location. Notably, HelixProtX outperforms BioMedGPT in three of these categories, indicating its superior accuracy in predicting diverse functional entities.

\subsection{Sequence Design}
\begin{figure*}[ht!]
\center
 \includegraphics[width=1.0\textwidth]{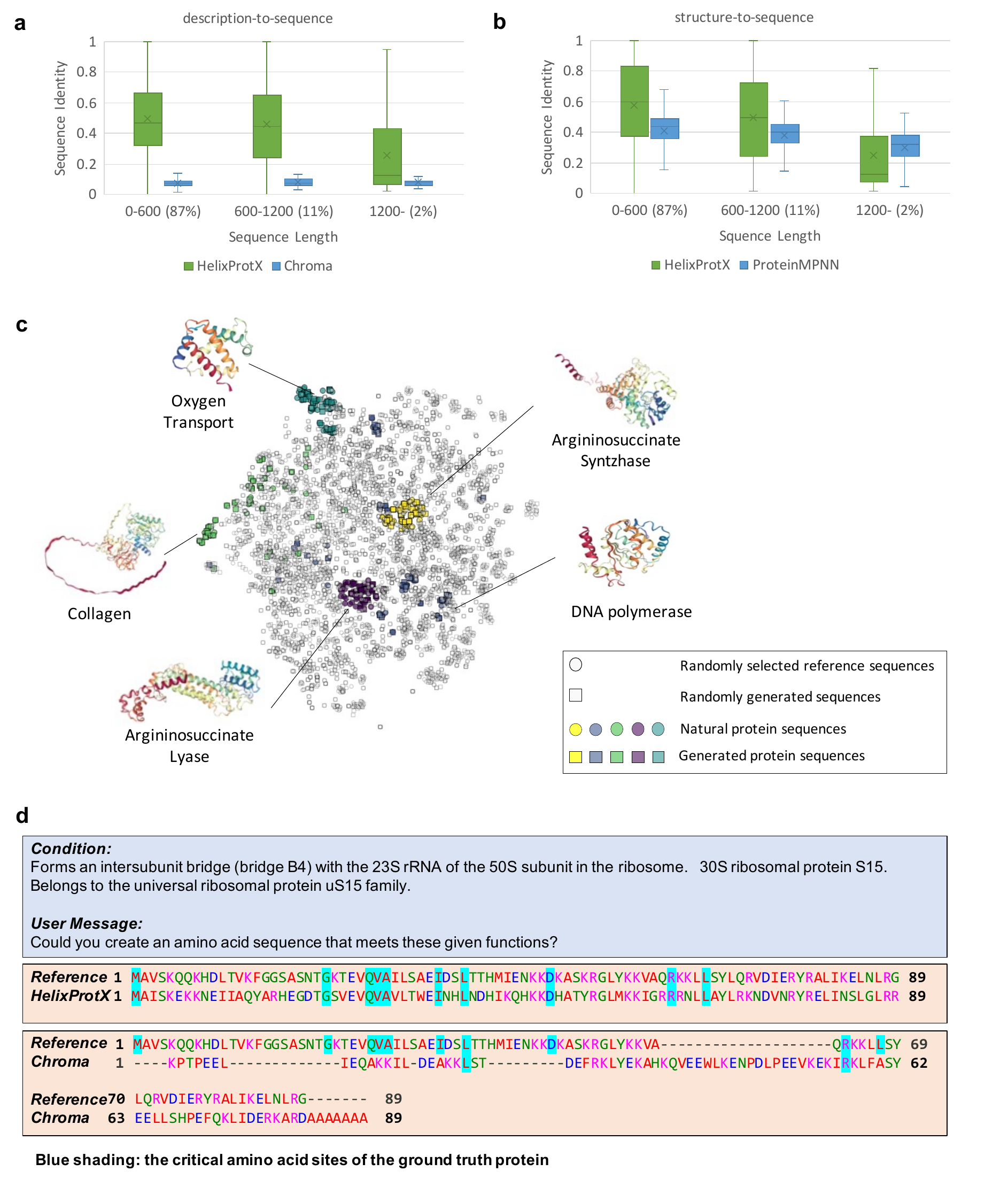}
\caption{\textbf{Results of sequence design.}
a. Performance comparison for description-to-sequence task.
b. Performance comparison for structure-to-sequence task.
c. Distribution of the reference sequences and designed sequence produced by HelixProtX for the description-to-sequence task.
d. An example of sequence design task.
}
\label{fig:sequence_design}

\end{figure*}

The protein sequence design task involves designing amino acid sequences with specific functions or structures based on input descriptions of desired functionalities or protein scaffold structures. This task has significant and wide-ranging applications.
\zhou{To evaluate the efficacy of HelixProtX, we compared it against two established methods}: ProteinMPNN \cite{prompnn}, which infers amino acid sequences from protein structures, and Chroma \cite{singh2024chroma}, which designs protein sequences based on textual descriptions. We use Sequence Identity \cite{steinegger2017mmseqs2} as the metric to \zhou{measure} the similarity between the sequences predicted by the model and the reference sequences.

\zhou{Initially}, we evaluate the overall accuracy of HelixProtX in the description-to-sequence task. As shown in Figure~\ref{fig:sequence_design}a, the protein sequences designed by HelixProtX \zhou{exhibit} significantly higher sequence identity with the reference sequences, outperforming the baseline method Chroma by several folds. This exceptional matching performance is particularly evident with shorter protein sequences. \zhou{Compared to the baseline method, We think this notable performance advantage is largely attributed to HelixProtX's utilization of a large-scale language model, which significantly enhances its capacity to comprehend and process textual descriptions.} 

In addition to evaluating the sequence similarity between the designed sequences and the reference sequences, we also assessed the \zhou{validity and diversity} of \zhou{the} designed \zhou{protein} sequences. Figure~\ref{fig:sequence_design}c \zhou{illustrates} the distribution of 3000 reference protein sequences and designed protein sequences for arbitrary text descriptions, represented by gray circles and squares, respectively. \zhou{Distributions for sequences designed for five key functions}: Argininosuccinate Lyase, RNA Polymerases, Argininosuccinate Synthase, Lactase, and Oxygen Transport, with 40 sequences per function, \zhou{are shown with colored circle and square markers.} We apply \zhou{ t-Distributed Stochastic Neighbor Embedding (t-SNE) to} visualize the data on a two-dimensional plane. \zhou{At first, this visualization demonstrates the validity of  HelixProtX. The analysis reveals that the overall distribution of the designed protein sequences closely matches that of the reference sequences. Specifically, the distribution of gray squares aligns well with the distribution of gray circles, affirming the plausibility of the designed sequences. Furthermore, the distribution of the designed protein sequences for each specific function closely matches that of the reference sequences. Second, this analysis also illustrates the diversity of these designed sequences. As we can see from Figure~\ref{fig:sequence_design}c, the designed protein sequences for each function exhibit considerable diversity, clustering in a relatively close area without collapsing into a single point, thereby demonstrating the model’s capacity to capture the inherent functional diversity inherent of protein sequences.}

In the structure-to-sequence task, HelixProtX \zhou{achieves} a notably high level of similarity between its designed sequences and the reference sequences, as \zhou{demonstrated} in Figure~\ref{fig:sequence_design}b. Across the vast majority (98\%) of samples, HelixProtX significantly surpasses ProteinMPNN, a baseline method specialized for inverse folding, in terms of sequence similarity. \zhou{Furthermore}, we utilize the HelixFold-Single model to predict the three-dimensional structures of the designed sequences and \zhou{assess} their alignment with the input structures using structure alignment scores. The results, \zhou{detailed in Table~\ref{tab:sequence_design_s} in Appendix,} further \zhou{confirm} the superiority of sequences designed by HelixProtX. 
Comparative analysis of the results of HelixProtX in Figure~\ref{fig:sequence_design}a and Figure~\ref{fig:sequence_design}b indicates that HelixProtX exhibits better performance in the structure-to-sequence task \zhou{than in} the description-to-sequence task. \zhou{This difference is likely due to the more definitive guidance provided by structural inputs compared to the inherent ambiguity in textual descriptions.}


In Figure~\ref{fig:sequence_design}d, we showcase a sequence designed by HelixProtX to assess the rationality of its amino acid sequence generation. To identify evolutionarily conserved functional key sites within the reference sequences, we use ESM-Scan\cite{Totaro2023.12.12.571273}, a tool based on the protein language model ESM-2\cite{Meier2021.07.09.450648}, \zhou{to identify key sites}. We then compare how well the sequences designed by HelixProtX and Chroma align with these identified key sites. Our analysis reveals that HelixProtX preserves nearly all of the critical sites found in the reference sequences, demonstrating its effectiveness in maintaining essential functional and structural elements. In contrast, the Chroma-designed sequences exhibit less consistency with these key sites. This high level of site preservation underscores HelixProtX's robustness and accuracy in protein design, validating its potential as a valuable tool for protein engineering.


\subsection{Structure Prediction / Design}

\begin{figure*}[ht]
\label{fig:structure_prediction}
\center
\includegraphics[width=1.0\textwidth]{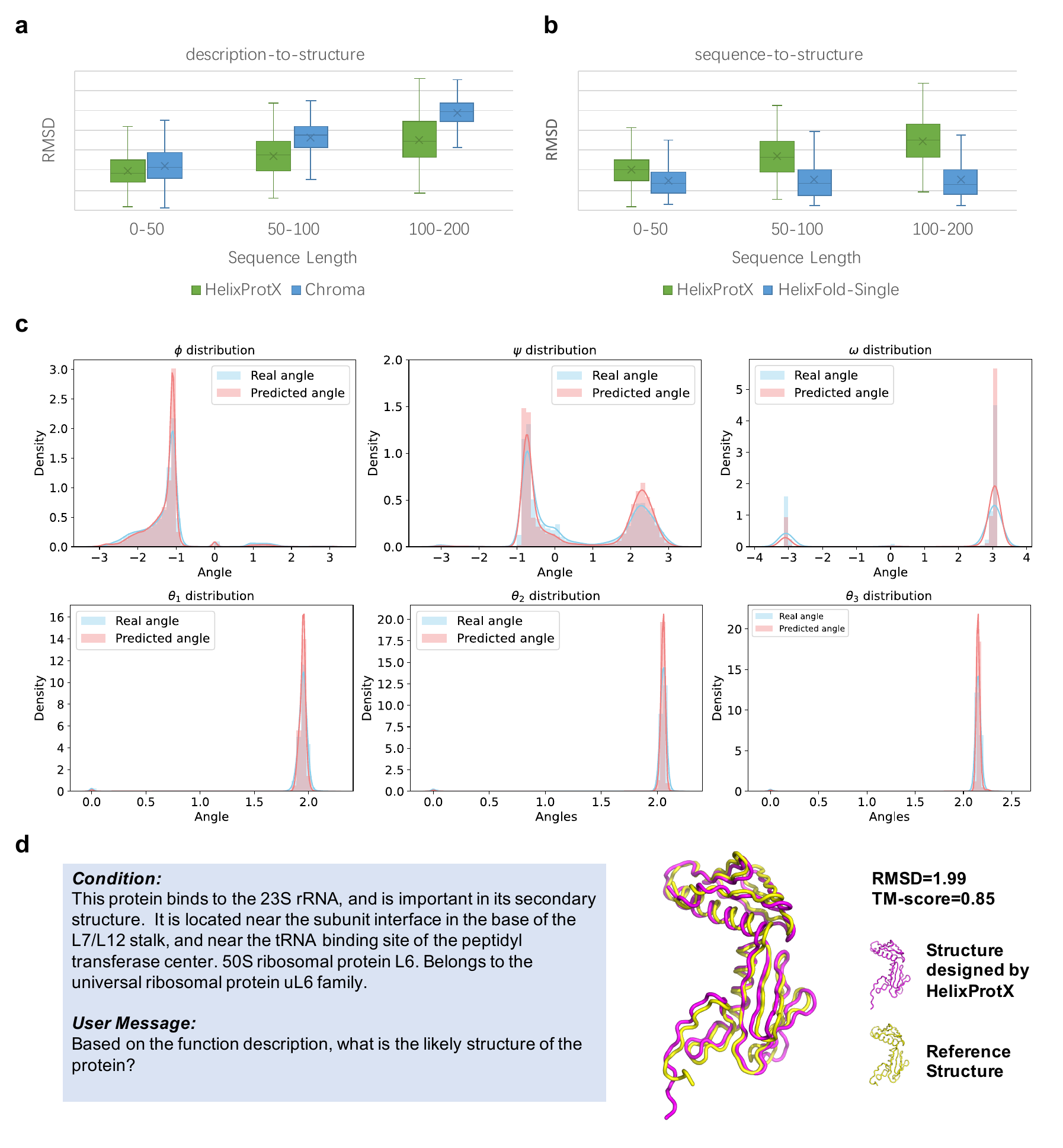}
\caption{\textbf{Results of structure prediction.}
a. Performance comparison for the description-to-structure task.
b. Performance comparison for the sequence-to-structure task.
c. The distribution of predicted protein angles versus actual protein angles.
d. An example of the description-to-structure task.
}

\label{fig:structure_prediction}
\end{figure*}

HelixProtX can not only predict the protein scaffold structure based on a given amino acid sequence (sequence-to-structure) but also design protein scaffold structures that meet specific functional requirements based on textual descriptions of protein functions (description-to-structure).
For the description-to-structure task, we once again use Chroma as our comparative baseline. As for the sequence-to-structure task, we compare HelixProtX with HelixFold-Single, which is \zhou{a well-established model for protein structure prediction } based on a protein language model. Root Mean Square Deviation (RMSD) is used to evaluate the error between the model-generated structures and the reference structures.


Figure~\ref{fig:structure_prediction}a compares the RMSD scores \zhou{between} HelixProtX and Chroma for the description-to-structure task. HelixProtX performs well across proteins of varying lengths, demonstrating its ability to effectively design protein backbones based on textual descriptions of their functions. \zhou{This capability is largely due to} the foundation model trained on a large corpus of text data, enabling it to effectively understand textual descriptions of proteins. Besides, as protein length increases, the spatial complexity of the design expands accordingly. Consequently, errors in model structure design tend to amplify. Figure~\ref{fig:structure_prediction}d presents an example of HelixProtX's performance on the description-to-structure task. The structure designed by HelixProtX closely matches the reference structure.


\zhou{In contrast to its performance in the description-to-structure task, HelixProtX exhibits less accurate results than HelixFold-Single in the sequence-to-structure task, also commonly referred to as protein folding. As shown in Figure~\ref{fig:structure_prediction}b, the prediction performance of HelixProtX, as measured by the Root Mean Square Deviation (RMSD), is notably less precise. This discrepancy can be attributed to several factors. First, HelixProtX faces challenges in accurately modeling the spatial relationships of amino acids. Unlike HelixFold-Single, which includes a specially designed component for superior geometric learning, HelixProtX uses a general language model approach that may not capture these complex spatial interactions effectively. Second, HelixProtX's method of decoding structural angles individually can lead to cumulative errors. This approach follows the common paradigm of LLMs but does not adequately address the complexities of protein structure prediction. Lastly, the process of transforming sequences into structures involves learning the complex physicochemical process of protein folding, a task of which general language models may have limited knowledge. This highlights the necessity for specially designed model architectures and training methodologies that are tailored to the challenges of unique tasks like sequence-to-structure transformation.}

HelixProtX \zhou{estimates} protein backbone structures by \zhou{predicting} six torsion angles of amino acids. We examined whether the distributions of these predicted angles align with the distributions of the actual angles. As shown in Figure~\ref{fig:structure_prediction}c, our observations indicate that the generated distributions closely mimic the real distributions, with almost complete overlap for all angles. This holds true for both the low-variance, near-Gaussian distributed angles ($\omega$, $\theta_{1}$, $\theta_{2}$, $\theta_{3}$) and the high-complexity distributed angles ($\phi$ and $\psi$).
\zhou{Such results provide indirect validation of the reliability and accuracy of the structural models generated by HelixProtX, proving its effectiveness in detailed protein structure prediction.}



\subsection{Advantages of Training a Unified Model for Protein-related Tasks}
\begin{figure*}[h]
\center
\includegraphics[width=0.7\textwidth]{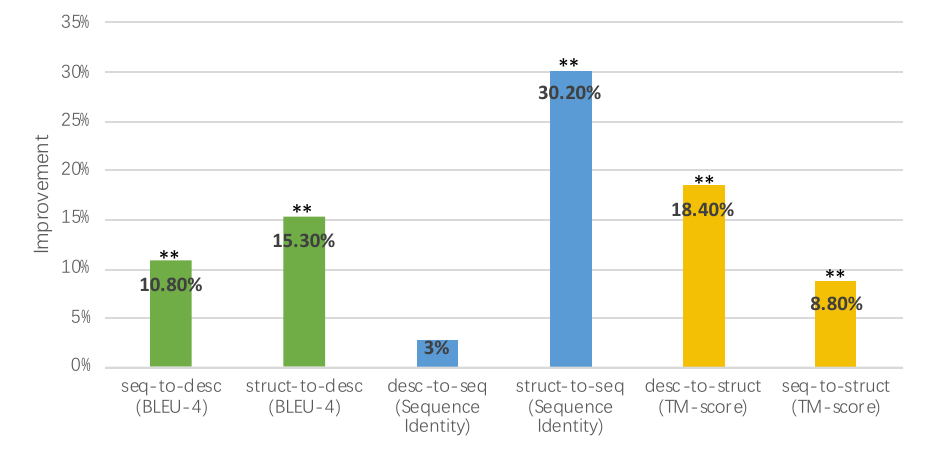}
\caption{Enhancement of joint training across various tasks in a unified model over applying task-specific training for each task. ``**'' indicates a significant improvement with a p-value less than 0.05.}
\label{fig:joint_training}
\end{figure*}

Most previous work employs different models to independently train each protein-related task. In contrast, HelixProtX utilizes a unified model to jointly train multiple protein-related tasks simultaneously. 
Figure~\ref{fig:joint_training}  examines the accuracy \zhou{benefits} of joint training multiple tasks within a unified model compared to independently training multiple task-specific models. \zhou{We employ} BLEU-4, Sequence Identity, and TM-score to evaluate \zhou{performance in} description prediction, sequence design, and structure prediction/design tasks, respectively. 

\zhou{Our results indicate that} joint training across various tasks in a unified model \zhou{yields} consistent improvements in accuracy over independent training for most tasks. This enhancement can be \zhou{largely} attributed to the model's ability to capture and \zhou{leverage} shared patterns and dependencies across tasks, leading to more robust and generalizable representations. The improvements are particularly notable in tasks involving structural modalities, whether these modalities serve as conditions or outputs. This is likely because the differences between structural modalities and text in large language models are substantial, and joint training effectively aligns these modalities, leveraging the strengths of large language models more efficiently.

\section{Conclusions and Future Work}

\zhou{

Proteins, as the fundamental building blocks of biological systems, can be represented through different modalities, including sequences, structures, and textual descriptions. Understanding the multimodal nature of protein data is crucial for scientific discoveries. HelixProtX, as a unified large multimodal model system, has successfully achieved the inter-mapping of different protein modalities. Our research results have demonstrated the superiority of HelixProtX in protein description prediction and sequence design tasks over baseline methods, validating the feasibility of HelixProtX for protein structure prediction/design tasks. Furthermore, our study has shown that employing a unified model for various protein-related tasks enhances the accuracy across multiple tasks. In summary, our study confirms the significant potential of large multimodal models in life sciences, offering new avenues for understanding protein biology and accelerating scientific discoveries in this domain. Promising directions for future research include:
\begin{itemize}
    \item Enhancing structure prediction/design: While HelixProtX shows promise, its performance in the structure prediction/design task requires further improvement.  Exploring more sophisticated network architectures and refining optimization objectives to more effectively capture the complexities of protein spatial structures may be possible for improving future task performance.
    \item Broadening the scope of applications: It is interesting to expand the dataset to include additional life science domains, such as small molecules and RNA. This can help to construct a large multimodal model with a broader application range, capable of handling diverse biological data types. 
\end{itemize}

}

\textbf{Code availability} 
The source code and inference code of HelixProtX are freely available on GitHub (https://github.com/PaddlePaddle/PaddleHelix/tree/dev/apps/helixprotx). 

\section*{Acknowledgments}
This work is supported by National Engineering Research Center of Deep Learning Technology and Applications and Ministry of Industry and Information Technology of the People's Republic of China (No.2021016612).

\clearpage

\begin{appendices}
\renewcommand{\thesection}{\arabic{section}}

\counterwithin{figure}{section}
\counterwithin{table}{section}
\counterwithin{equation}{section}

\section{Dataset}
We have meticulously curated a comprehensive multimodal protein dataset by integrating resources from UniProtQA \cite{luo2023biomedgpt} and SwissProt \cite{boeckmann2003swiss}. From UniProtQA, we extracted amino acid sequences and their corresponding functional descriptions, all derived from the extensive UniProt database \cite{uniprot2015uniprot}. Additionally, we obtained amino acid sequences and their corresponding three-dimensional structures from SwissProt. 
By matching the amino acid sequences, we paired the functional descriptions from UniProtQA with the structures from SwissProt. This integration has resulted in the creation of a multimodal dataset comprising 361,498 proteins.
Each protein's information can be used to generate six samples, corresponding to the tasks: sequence-to-function, structure-to-function, description-to-sequence, structure-to-sequence, description-to-structure, and sequence-to-structure. In addition to the multimodal protein data, we augmented our training with self-collected conversational data, ensuring the model maintains strong conversational capabilities. Furthermore, we applied a filtration criterion to exclude samples with a token count exceeding 1024, streamlining the dataset for efficient model training.

\section{Model Architecture}

HelixProtX takes three inputs: the system message $\mathbf{m}^\text{sys}$, the user message $\mathbf{m}^\text{usr}$, and the condition $\mathbf{f}^\text{cond}$. Both $\mathbf{m}^\text{sys}$ and $\mathbf{m}^\text{usr}$ are sequences of textual tokens. The condition $\mathbf{f}^\text{cond}$ can be one of the following: the text token sequence $\mathbf{f}^\text{desc}$ representing the protein description, the amino acid sequence $\mathbf{f}^\text{seq}$, or the coordinates of all atoms in the protein $\mathbf{f}^\text{struct}$. For description prediction and sequence design tasks, HelixProtX generates a textual response $\mathbf{X}$. For structure prediction and design tasks, HelixProtX produces a sequence of residue angles $\mathbf{A}$. 

Algorithm \ref{algorithm:inference} outlines the main inference steps of HelixProtX. 
Lines 1 to 7 organize and concatenate the input information to form the LLM input $\mathbf{f}^\text{input}$. The sequence condition, structure condition, and textual description condition are each processed through their respective Sequence Encoder, Structure Encoder, and the LLM's WordEmbedding for feature transformation.
Lines 9 to 18 are focused on decoding outputs based on the specific task. For description prediction and sequence design tasks, text tokens are decoded by the LanguageModelHead (line 11). For structure prediction and design tasks, residue angles are decoded by the ResidueAngleHead (line 15) when the predicted text token is [NUM]. $\mathbf{token}^\text{EOS}$ and $\mathbf{token}^\text{NUM}$ denote the embedding of specific tokens [EOS] and [NUM], respectively. $d_h$=4096 is the hidden size of the language model. Both the LanguageModelHead and the ResidueAngleHead are implemented as MLP layers. The decoded residue angles are then processed by the ResidueAngleEmbedding module (see Subsection \ref{subsec:angle_emb}) to prepare them for subsequent prediction iterations.

\begin{algorithm}
\caption{HelixProtX Model Inference}
\label{algorithm:inference}
\KwIn{$\mathbf{m}^\text{sys}$, $\mathbf{m}^\text{usr}$, $\mathbf{f}^\text{cond}=\{\mathbf{f}^\text{seq}, \mathbf{f}^\text{struct}, \mathbf{f}^\text{desc} \}$}
\KwOut{$\mathbf{X}$, $\mathbf{A}$}

$\mathbf{f}^\text{input} = \text{WordEmbedding}(\mathbf{m}^\text{sys}+\mathbf{m}^\text{usr})$

\If{$\mathbf{f}^\text{seq}$ is not None}{

    $\mathbf{f}^\text{input} = \text{Concat}(\mathbf{f}^\text{input}, \text{Abstractor}(\text{SequenceEncoder}(\mathbf{f}^\text{seq}))$
    
}

\If{$\mathbf{f}^\text{struct}$ is not None}{
    $\mathbf{f}^\text{input} = \text{Concat}(\mathbf{f}^\text{input}, \text{Abstractor}(\text{StructureEncoder}(\mathbf{f}^\text{struct})))$
    
}

\If{$\mathbf{f}^\text{desc}$ is not None}{
$\mathbf{f}^\text{input} = \text{Concat}(\mathbf{f}^\text{input}, \text{WordEmbedding}(\mathbf{f}^\text{desc}))$

}

$\mathbf{A}, \mathbf{X} = \emptyset, \emptyset$\;

\While{$\mathbf{x}^\text{next}!=\mathbf{token}^\text{EOS}$}{

    $\mathbf{h} = \text{LanguageModel}( \mathbf{f}^\text{input})$ \hfill
    $\mathbf{h} \in   \mathbb{R}^{d_{\mathrm{h}}}$
    
    $\mathbf{x}^\text{next} = \text{LanguageModelHead}(\mathbf{h})$
    
    $\mathbf{X} = \text{Concat}(\mathbf{X},\mathbf{x}^\text{next})$
    
    $\mathbf{f}^\text{next} = \text{WordEmbedding}(\mathbf{x}^\text{next})$

    \If{$\mathbf{x}^\text{next}=\mathbf{token}^\text{NUM}$}{
        $\mathbf{a}^\text{next} = \text{ResidueAngleHead}(\mathbf{h})$ \hfill  
        $\mathbf{a}^\text{next} \in   \mathbb{R}^6 $ 
        
        $\mathbf{A} = \text{Concat} (\mathbf{A},\mathbf{a}^\text{next})$ 
        
        $\mathbf{f}^\text{next} += \text{ResidueAngleEmbedding}(\mathbf{a}^\text{next})$
        
    }
    $\mathbf{f}^\text{input} = \text{Concat}(\mathbf{f}^\text{input}, \mathbf{f}^\text{next})$
    
}
\Return $\mathbf{X}$, $\mathbf{A}$

\end{algorithm}

\begin{algorithm}
\caption{Abstractor}
\label{algorithm:abstractor}
\KwIn{$\mathbf{f}^\text{enc} $}
\KwOut{$\mathbf{f}^\text{abs}$}

\For{$\textbf{all } i \in [1, \dots, N_\text{layers}]$}{
    $\mathbf{Q} = \text{LayerNorm}(\text{Linear}(\mathbf{t^{abs}}))$ \hfill
    $\mathbf{Q} \in   \mathbb{R}^{ n_{\mathrm{abs}}   \cdot d_{h}^{\mathrm{abs}} }$
    
    $\mathbf{K}, \mathbf{V} = \text{LayerNorm}(\text{Linear}(\mathbf{f}^\text{enc})), \text{LayerNorm}(\text{Linear}(\mathbf{f}^\text{enc}))$ \hfill 
    $\mathbf{K}, \mathbf{V} \in   \mathbb{R}^{ n_{\mathrm{r}}  \cdot d_{h}^{\mathrm{abs}} }$
    
    $\mathbf{O} = \text{Attention}(\mathbf{Q},\mathbf{K},\mathbf{V})$ \hfill 
    $\mathbf{O} \in   \mathbb{R}^{n_{\mathrm{abs}}  \cdot d_{h}^{\mathrm{abs}} } $
    
    $\mathbf{f}^\text{abs} +=  \text{Linear}(\mathbf{O})$ \hfill
    $\mathbf{f}^\text{abs} \in   \mathbb{R}^{  n_{\mathrm{abs}}  \cdot d_{h}^{\mathrm{abs}} }$
    
    $\mathbf{f}^\text{abs} +=  \text{MLP}(\text{LayerNorm}(\mathbf{f}^\text{abs}))$
    
}
$\mathbf{f}^\text{abs} \gets \text{Linear}(\mathbf{f}^\text{abs})$ \hfill
    $\mathbf{f}^\text{abs} \in   \mathbb{R}^{n_{\mathrm{abs}}    \cdot d_{\mathrm{h}} }$


\Return $\mathbf{f}^\text{abs}$
\end{algorithm}

\begin{algorithm}
\caption{Residue Angle Embedding}
\label{algorithm:angle_embed}
\KwIn{$\mathbf{a}_{i}  $}

\KwOut{$\mathbf{f}^\text{angle}$}

$\mathbf{f}^\text{angle} = \emptyset$

\For{$\textbf{all } j \in [1, \dots, n_{\mathrm{a}}]$}{
    $\mathbf{a}_{ij} \gets(\mathbf{a}_{ij} +\pi)\mod{2\pi}-\pi$ 

    $\mathbf{k} = \text{GetIndex}(\mathbf{a}_{ij}, [-\pi, \pi), N_b)$
    
    $\mathbf{f} = \frac{N_{b}}{2\pi}((\mathbf{a}_{ij}- \mathbf{a}_k^{bin}) \cdot \mathbf{f}_k^{base} + (\mathbf{a}_{k+1}^{bin}-\mathbf{a}_{ij})\cdot \mathbf{f}_{k+1}^{base})$

    $\mathbf{f}^\text{angle} = \text{Concat}(\mathbf{f}^\text{angle} , \mathbf{f})$

}

$\mathbf{Q},\mathbf{K},\mathbf{V} = \text{Linear}(\mathbf{f}^\text{angle})$ \hfill $\mathbf{Q},\mathbf{K},\mathbf{V} \in   \mathbb{R}^{n_{\mathrm{a}}   \cdot d_{\mathrm{h}}} $

$\mathbf{O} = \text{Attention}(\mathbf{Q},\mathbf{K},\mathbf{V})$ \hfill
$\mathbf{O} \in   \mathbb{R}^{n_{\mathrm{a}}    \cdot d_{\mathrm{h}} } $

$\mathbf{f}^\text{angle} = \text{MLP}(\mathbf{O})$

\Return $\mathbf{f}^\text{angle}$
\end{algorithm}




    


\subsection{Abstractor}
\label{subsec:abstractor}
Abstractor modules (Algorithm \ref{algorithm:abstractor}) map sequence and structure representations into the LLM's input space. We simplify the Abstractor designed by \cite{ye2023mplugowl}, removing the concatenation operation between the trainable token $\mathbf{t^{abs}}$ and the input vector $\mathbf{f}^\text{enc}$.

The Abstractor module takes the output of the Sequence Encoder $\mathbf{f}_{seq}^\text{enc} \in \mathbb{R}^{n_{\mathrm{r}} \cdot d_{\mathrm{enc}}}$ with a dimensionality of $d_{\mathrm{enc}}=384$, or the output of the Sequence Encoder $\mathbf{f}_{struct}^\text{enc} \in \mathbb{R}^{n_{\mathrm{r}} \cdot d_{\mathrm{enc}}}$  with a dimensionality of $d_{\mathrm{enc}}=128$ as input, and outputs the transformed feature $\mathbf{f}^\text{abs} \in \mathbb{R}^{n_{\mathrm{abs}} \cdot d_{\mathrm{h}} }$, where $n_r$ is the residue number of the protein, $d_{\mathrm{h}}$ is the LLM embedding dimension, and $n_{\mathrm{abs}}=64$ is the number of tokens after abstraction.

The trainable tokens $\mathbf{t}^{\mathrm{abs}} \in \mathbb{R}^{n_{\mathrm{abs}} \cdot d_{\mathrm{enc}}}$, introduced in \cite{ye2023mplugowl}, are a set of learnable parameters utilized to align the features from the sequence/structure encoder with those of the LLM via a cross-attention mechanism. In this mechanism, we designate either $f_{seq}^\text{enc}$ or $f_{struct}^\text{enc}$ as the key and value, while the $\mathbf{t^{abs}}$ within the abstractors serve as the query (Line 3). Additionally, the hidden dimension $d_{h}^{\mathrm{abs}}$ in the abstractor is set to 32.

\subsection{Residue Angle Embedding}
\label{subsec:angle_emb}
The Residue Angle Embedding (Algorithm \ref{algorithm:angle_embed}) maps the embedding space of the residue angles to the embedding space of the text.  It takes the residue angles $\mathbf{a}_i \in \mathbb{R}^{ n_{\mathrm{a}} } $ as input and produces the embedding angle $\mathbf{f}^\text{angle} \in \mathbb{R}^{n_{\mathrm{a}} \cdot d_{\mathrm{h}}}$, where $n_{\mathrm{a}}=6$ and angle $\mathbf{a}_{ij} \in [-\pi, \pi)$.
The interval $[-\pi, \pi)$ is divided into $N_{\mathrm{b}}$ equal segments, denoted as $\{[\mathbf{a}_k^{bin},\mathbf{a}_{k+1}^{bin})\}|_{k=0}^{N_b-1}$, where $N_{\mathrm{b}}=64$. Each bin corresponds to a learnable base embedding, denoted as $\mathbf{f}^{base}_k\in \mathbb{R}^{d_{\mathrm{b}}}$, with $d_{\mathrm{b}}=128$. 

We use a linear interpolation method to calculate the embedding for each residue's angle $\mathbf{a}_{ij}$. Specifically, we first identify the interval in which the angle $\mathbf{a}_{ij}$ falls (assuming that it is the $k$-th interval) through $\text{GetIndex}(\mathbf{a}_{ij}, [-\pi, \pi), N_b)$. Then we compute the embedding by weighting the corresponding base embeddings $f_{k}^{base}$ and $f_{k+1}^{base}$ based on the distance of the angle value from the segment endpoints $a_k^{bin}$ and $a_{k+1}^{bin}$ of the interval $k$. Then, this angle's embedding is transformed through self-attention to obtain the final output.

\section{Settings of Training and Evaluation}

\subsection{Training Settings}
HelixProtX employs a two-stage training approach. Stage 1 involves training for 2 epochs with a learning rate of $5 \times 10^{-4}$. Stage 2 follows with 50 epochs with a reduced learning rate of $5 \times 10^{-5}$. During training, the model processes 102,400 tokens per batch. The experiments are conducted using PaddlePaddle \cite{ma2019paddlepaddle}, leveraging 8 NVIDIA A100 GPUs, each with 40 GB of memory. To improve training efficiency, we employ the Zero Redundancy Optimizer (ZeRO) \cite{10.5555/3433701.3433727} and use the bf16 data format.




\subsection{Evaluation Metrics}
The Root Mean Square Deviation (RMSD) is a metric used to measure the distance between the spatial structures of two proteins. It is defined as the square root of the average of the squared differences in coordinates between corresponding atoms in the aligned protein structures.

TM-score \cite{zhang2004scoring} ranges between 0 and 1, which represents the similarity of the alignment structure. A higher TM-score value usually indicates a higher similarity between compared structures. 

The F1 Score \cite{berners1994world} evaluates classification performance, particularly with imbalanced datasets, by combining precision and recall into a single metric. It is the harmonic mean of these two metrics, offering a balanced measure when precision and recall differ significantly.

Bilingual Evaluation Understudy\cite{papineni2002bleu} (BLEU) is a metric used to automatically evaluate the quality of machine translation, with the core idea of comparing the n-gram similarity between the output of machine translation and the standard translation.

Recall-Oriented Understudy for Gisting Evaluation\cite{lin2004rouge} (ROUGE) is a common evaluation metric used in fields such as machine translation, automatic summarization, and question answering. It assesses translation quality by counting the number of units that overlap (such as n-grams, word sequences, and word pairs) between the statistical translation output and the standard reference.

Sequence identity refers to the ratio of identical amino acids at the corresponding positions in two aligned sequences to the length of the longer sequence. We calculate it using MMseqs2\cite{steinegger2017mmseqs2}.

\subsection{Baseline Methods}
We compare HelixProtX with several state-of-the-art baseline methods designed for various protein-related tasks. Detailed descriptions of these baseline methods are provided in Table~\ref{tab:basline_model}.

\begin{table}[h]
\centering
\caption{Baseline methods}
\label{tab:basline_model}
\begin{tabular}{|c|c|>{\centering\arraybackslash}m{10cm}|}  
\hline
Method & Task & Method introduction \\ \hline

BioMedGPT\cite{luo2023biomedgpt} & seq-to-desc & BioMedGPT is an open multimodal generative pre-trained transformer (GPT) for biomedicine that unifies the feature spaces of molecules, proteins, and natural language via encoding and alignment. It excels in protein domain question-answering tasks by accurately interpreting protein sequences. \\ \hline

ProteinMPNN\cite{prompnn} & seq-to-struct & The ProteinMPNN model is trained on a curated PDB dataset and refined with backbone noise, which advances sequence design capabilities. \\ \hline

Chroma\cite{singh2024chroma} & \makecell{desc-to-struct\\desc-to-seq} & Chroma is a protein generation model that introduces a diffusion process for the conformational statistics of polymer aggregates, allowing for the direct sampling of new protein structures and sequences, which can be tuned to steer the generation process toward desired properties and functions. \\ \hline

HelixFold-Single\cite{fang2023method} & seq-to-struct & HeliXFold-Single is a single sequence-based protein structure prediction model. The model extracts information from nearly 300 million unlabeled protein data, models relationships between proteins, and thus effectively replaces the MSA information retrieval module, greatly improving the speed of structure prediction. \\ \hline

\end{tabular}
\end{table}

\section{Supplemental Results}
\subsection{Self-Consistency}

\begin{figure*}[h]
\center
\includegraphics[width=0.9\textwidth]{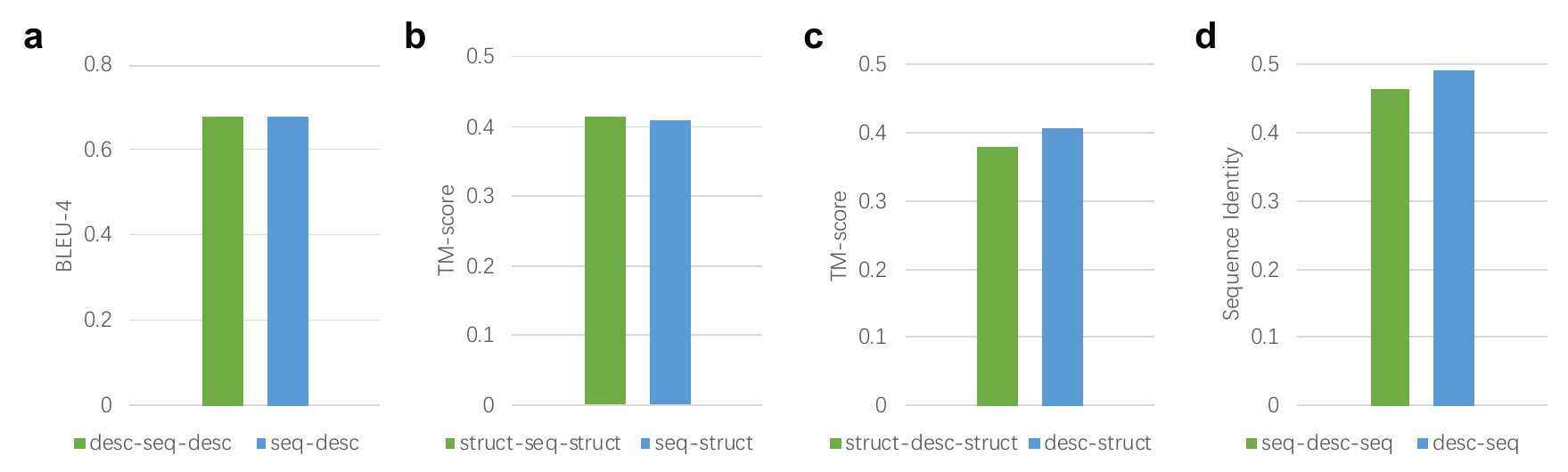}
\caption{Self-consistency of HelixProtX.
}

\label{fig:self-consistency}
\end{figure*}
The self-consistency of a model can be an important indicator of its reliability. To evaluate the self-consistency of HelixProtX, we adapted a methodology inspired by the Back Translation Test \cite{edunov2018understanding}. This test typically involves translating output from a machine translation model back into the source language and comparing it to the original input. Similarly, we treat the sequence, structure, and description modalities of proteins as distinct ``languages'' to assess the consistency of the model's outputs.  We compare the results of HelixProtX when using the input $x$ of modality $X$ as the condition to predict the output $\hat{y}$ of modality $Y$, and then using $\hat{y}$ as the condition to predict the output $\hat{x}$ of modality $X$, denoted as X-Y-X, with the results obtained when HelixProtX directly predicts $\hat{x}$ of modality $X$ from modality $Y$ (denoted as Y-X).

The self-consistency results of HelixProtX are illustrated in Figure~\ref{fig:self-consistency}. As HelixProtX is limited to predicting only the backbone of protein structures, we were unable to evaluate scenarios where Y is the structure modality. Despite this, the results for all X-Y-X and Y-X pairs are closely aligned, reflecting the high self-consistency of HelixProtX.

\subsection{Sequence Design}

\begin{table}[h]
\centering
\caption{Supplemental results for sequence design}
\label{tab:sequence_design_s}
\begin{tabular}{c|c|c|c|c}
\hline
                                  & Task & Sequence Identity                                                                & RMSD & TM-score   \\ \hline
MPNN                      & struct-to-seq  &         0.401   &   2.426   &         0.787                       \\ \hline
Chroma                      & desc-to-seq  &         0.075   &   6.137   &         0.284                      \\ \hline
\multirow{2}{*}{HelixProtX}       & struct-to-seq   &         0.564   &   2.257   &       0.792                     \\   \cline{2-5} 
                                  & desc-to-seq    &       0.486    &   2.415   &    0.770                         \\ \hline
\end{tabular}
\end{table}

In Table \ref{tab:sequence_design_s}, we present the comprehensive results of the sequence design task, detailing the performance across various metrics.

\subsection{Structure Prediction/Design}
In Table \ref{tab:structure_prediction_s}, we present the comprehensive results of the structure prediction/design task, detailing the performance across various metrics.

\begin{table}[h]
\centering
\caption{Supplemental results for structure prediction/design}
\label{tab:structure_prediction_s}
\begin{tabular}{c|c|c|c}
\hline
                            & Task                                                                 & RMSD & TM-score   \\ \hline
HelixFold-Single            & seq-to-struct       &    1.497  &         0.770                    \\ \hline
Chroma                      & desc-to-struct    &  3.558    &        0.308                 \\ \hline
\multirow{2}{*}{HelixProtX} & seq-to-struct       &    2.723  &        0.416                 \\ \cline{2-4} 
                            & desc-to-struct     &   2.724   &       0.407                 \\ \hline
\end{tabular}
\end{table}

\subsection{Advantage of Training a Unified Model}
In Table~\ref{tab:unified_and_independent}, we present a comprehensive comparison of the performance of joint training multiple tasks into a unified model versus independently training each task across six protein-related tasks.

\begin{table}[h]
\centering
\caption{Comparison of unified training and independent training}
\label{tab:unified_and_independent}

\begin{tabular}{>{\centering\arraybackslash}p{2.85cm}|>{\centering\arraybackslash}p{2.85cm}|>{\centering\arraybackslash}p{2.0cm}|>{\centering\arraybackslash}p{2.0cm}|>{\centering\arraybackslash}p{2.0cm}|>{\centering\arraybackslash}p{2.0cm}}
\hline
\multicolumn{1}{l|}{}                 & Task & BLEU-4 & ROUGE1-F1 & ROUGE2-F1 & ROUGEL-F1 \\ \hline
\multirow{2}{*}{Unified model}     &  struct-to-desc    &  0.624      &     0.829     &   0.703       &    0.826      \\ \cline{2-6} 
                                      &  seq-to-desc    &   0.644     &      0.850    &   0.725       &   0.847       \\ \hline
\multirow{2}{*}{Independent model} &  struct-to-desc    &   0.541     &      0.752    &     0.619     &    0.749      \\ \cline{2-6} 
                                      &  seq-to-desc    &    0.581    &     0.791     &    0.661      &     0.788     \\ \hline
\end{tabular}

\vspace{0.1cm} 

\begin{tabular}{>{\centering\arraybackslash}p{2.85cm}|>{\centering\arraybackslash}p{2.85cm}|>{\centering\arraybackslash}p{2.66cm}|>{\centering\arraybackslash}p{2.67cm}|>{\centering\arraybackslash}p{2.66cm}}
\hline
\multicolumn{1}{l|}{}                 & Task                     & RMSD & TM-score   & Sequence Identity \\ \hline
\multirow{2}{*}{Unified model}     & struct-to-seq      &  3.335    &      0.643                           & 0.362\\ \cline{2-5} 
                                      & desc-to-seq    &  3.353   &      0.638                         & 0.354  \\ \hline
\multirow{2}{*}{Independent model} & struct-to-seq      &  4.077    &     0.527                             & 0.278  \\ \cline{2-5} 
                                      & desc-to-seq     &   3.466   &     0.604                          & 0.344    \\ \hline
\end{tabular}

\vspace{0.1cm} 

\begin{tabular}{>{\centering\arraybackslash}p{2.85cm}|>{\centering\arraybackslash}p{2.85cm}|>{\centering\arraybackslash}p{4.1cm}|>{\centering\arraybackslash}p{4.1cm}}
\hline
\multicolumn{1}{l|}{}                 & Task                     & RMSD & TM-score  \\ \hline
\multirow{2}{*}{Unified model}     & desc-to-struct       &   2.993                &      0.316           \\ \cline{2-4}
                                      & seq-to-struct    &   2.939                &       0.333          \\ \hline
\multirow{2}{*}{Independent model} & desc-to-struct       &  3.250                  &        0.267         \\ \cline{2-4}
                                      & seq-to-struct     &  3.238                 &      0.306           \\ \hline
\end{tabular}

\end{table}

\end{appendices}

\clearpage

\bibliographystyle{unsrt}  
\bibliography{references}

\end{document}